\documentclass[letterpaper, 10 pt, journal, twoside]{ieeetran}
\usepackage{ieee_robotics}
\newcommand{\method}{\textit{MultiGrasp}\xspace}
\newcommand{\dataset}{\textit{Grasp'Em}\xspace}

\acrodef{ddpm}[DDPM]{Denoising Diffusion Probabilistic Model}
\acrodef{mala}[MALA]{Metropolis-Adjusted Langevin Algorithm}
\acrodef{sdf}[SDF]{Signed Distance Function}
\acrodef{fkine}[FK]{Forward Kinematics}
\acrodef{ik}[IK]{Inverse Kinematics}
\acrodef{ppo}[PPO]{Proximal Policy Optimization}
\acrodef{rl}[RL]{Reinforcement Learning}
\acrodef{gcrl}[GCRL]{Goal-Conditioned Reinforcement Learning}
\acrodef{mdp}[MDP]{Markov Decision Process}
\acrodef{ood}[OOD]{Out-of-Domain}
\acrodef{ibs}[IBS]{Intersection Bisector Surface}
\acrodef{bp}[BP]{Behavior-Paradigm}
\acrodef{dof}[DoF]{Degrees of Freedom}
\acrodef{dfc}[DFC]{Differentiable Force Closure}

\title{
Grasp Multiple Objects with One Hand
}

\begin{document}

\let\oldtwocolumn\twocolumn
\renewcommand\twocolumn[1][]{%
    \oldtwocolumn[{#1}{
        \vspace{-33pt}
        \centering
        \includegraphics[width=0.9\linewidth]{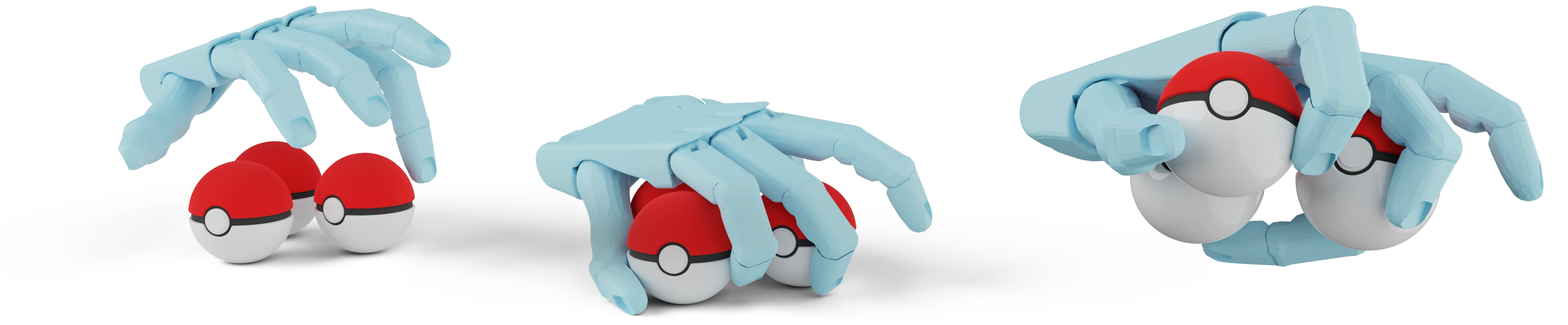}
        \captionof{figure}{\textbf{The proposed multi-object grasping method, \method, drives the Shadow Hand to simultaneously grasp multiple Pokémons.}}
        \label{fig:teaser}
        \vspace{3pt}
    }]
}

\author{Yuyang Li$^{1,2,3}$, Bo Liu$^{3}$, Yiran Geng$^{4}$, Puhao Li$^{1,2}$,Yaodong Yang$^{1,3}$, Yixin Zhu$^3$, Tengyu Liu$^{1}$, Siyuan Huang$^{1}$
\thanks{Manuscript received: November 22, 2023; Revised: January 28, 2024; Accepted: February 23, 2024.}%
\thanks{This letter was recommended for publication by Associate Editor N. Rojas and Editor J. Borrãs Sol upon evaluation of the reviewers' comments.}%
\thanks{This work is supported in part by the National Science and Technology Major Project (2022ZD0114900) and the Beijing Nova Program.}%
\thanks{$^\dagger$ Corresponding emails: {\tt{\{liutengyu, syhuang\}@bigai.ai}}. See additional material on \href{https://multigrasp.github.io}{https://multigrasp.github.io}.}%
\thanks{$^1$~National Key Laboratory of General AI, Beijing Institute for General Artificial Intelligence.
$^2$~Department of Automation, Tsinghua University.
$^3$~Institute for Artificial Intelligence, Peking University.
$^4$~School of Electronics Engineering and Computer Science, Peking University.}%
\thanks{Digital Object Identifier (DOI): 10.1109/LRA.2024.3374190}%
}%

\markboth{IEEE ROBOTICS AND AUTOMATION LETTERS. PREPRINT VERSION. ACCEPTED FEBRUARY, 2024}{Li \MakeLowercase{\etal}: Grasp Multiple Objects with One Hand}

\maketitle

\begin{abstract}
The intricate kinematics of the human hand enable simultaneous grasping and manipulation of multiple objects, essential for tasks such as object transfer and in-hand manipulation. Despite its significance, the domain of robotic multi-object grasping is relatively unexplored and presents notable challenges in kinematics, dynamics, and object configurations. This paper introduces \method, a novel two-stage approach for multi-object grasping using a dexterous multi-fingered robotic hand on a tabletop. The process consists of (i) generating pre-grasp proposals and (ii) executing the grasp and lifting the objects. Our experimental focus is primarily on dual-object grasping, achieving a success rate of 44.13\%, highlighting adaptability to new object configurations and tolerance for imprecise grasps. Additionally, the framework demonstrates the potential for grasping more than two objects at the cost of inference speed.
\end{abstract}


\section{Introduction}

\IEEEPARstart{I}{nfants}, between 6 to 9 months of age, transition from using a rudimentary ``grabbing'' technique with their entire hand to a more refined ``pincer grasp,'' involving only a subset of fingers~\cite{von2004action}. This developmental progression underpins advanced object manipulation skills, including the ability to grasp multiple objects~\cite{moll2010infant,billard2019trends}. Similarly, in robotics, significant advancements have been made in multi-fingered dexterous hands~\cite{xu2023unidexgrasp,wan2023unidexgrasp++,li2023gendexgrasp,liu2021synthesizing,she2022learning,yao2023exploiting}, enabling intricate grasping and in-hand manipulation tasks~\cite{mason1985robot,dafle2014extrinsic,rus1999hand,calli2018learning,andrychowicz2020learning,chen2022system,yin2023rotating} and enhancing interaction capabilities for embodied intelligence.

However, the majority of existing research in robotic grasping focuses on single-object scenarios~\cite{liu2021synthesizing,she2022learning,xu2023unidexgrasp,wan2023unidexgrasp++,li2023gendexgrasp}. Common strategies often mirror the action of enveloping the object with the hand and squeezing the fingers towards it~\cite{miller2004graspit,li2023gendexgrasp}, effectively reducing the use of sophisticated dexterous hands to mere parallel grippers. This overlooks the rich potential of their complex articulated structure and kinematic redundancy.

In this work, we delve into the less-explored domain of multi-object grasping, an intricate task that necessitates meticulous management of the hand's dexterous kinematics and dynamics. Our objective is to manipulate a multi-fingered dexterous hand to simultaneously grasp and lift multiple objects placed on a table. Distinct from single-object grasping, multi-object grasping demands independent force closure on each object. In this scenario, each object is a separate entity with no rigid interconnection, presenting unique challenges:

\paragraph*{Diverse Configurations}

Multi-object grasping encompasses a broad spectrum of object configurations influenced by varying geometries, combinations, and placements. This diversity is further compounded by the various hand configurations, necessitating the development of adaptable and flexible grasping strategies~\cite{billard2019trends,yao2023exploiting}.

\paragraph*{Intricate Kinematics}

The task of multi-object grasping requires using the full extent of the hand's workspace, as each object occupies a significant portion of it. Simple contacts via the palm or fingertips are insufficient. Instead, the entire length and sides of the fingers must be engaged~\cite{billard2019trends,yao2023exploiting}, necessitating a carefully configured grasping pose for effective force closure on each object while avoiding collisions.

\paragraph*{Complex Dynamics}

The traditional \textit{enveloping and squeezing} approach, typically employed in single-object grasping, is inadequate for multi-object scenarios. Repositioning a finger to better grasp one object might jeopardize the grasp on another. Therefore, precise control and fine-tuning of the wrenches at each contact point become imperative.

\begin{figure*}[t!]
    \centering
    \includegraphics[width=\linewidth]{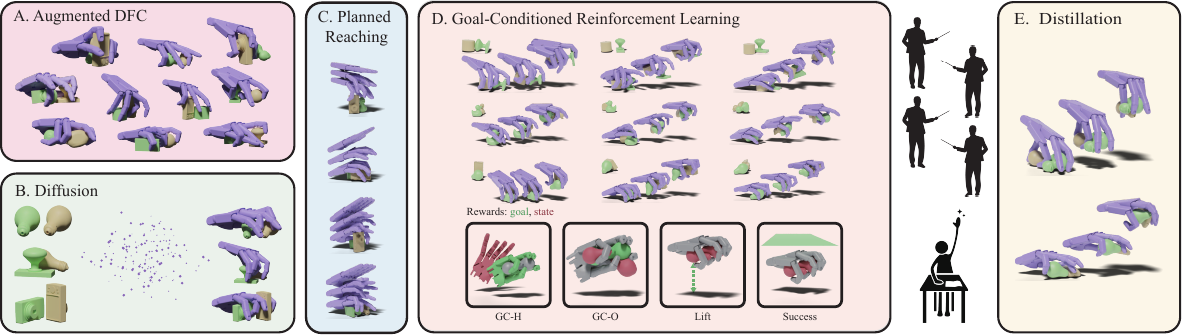}
    \caption{\textbf{Overview of \method.} The pre-grasp pose proposal module (A, B) generates an optimal hand pose to grasp the target objects. A motion planning module (C) then plans the reaching trajectory from a flat hand to the desired pose. For lifting, a suite of specialist \ac{rl} policies (D) are deployed, tailored for different object configurations. These policies are subsequently distilled (E) to develop a vision-based policy suitable for real-world application.}
    \label{fig:pipeline}
\end{figure*}

To address these challenges, we introduce \method, a computational framework devised for multi-object grasping. As depicted in \cref{fig:pipeline}, \method begins by generating a pre-grasp pose for the given target objects. This pose represents a preliminary hand configuration, serving as a goal strategy for execution. Utilizing this strategy, the framework employs an execution policy to control the hand in picking up the objects. To demonstrate this process, we have constructed \dataset, a large-scale synthetic dataset comprising 90k diverse multi-object grasps, utilizing the Shadow Hand. 
Moreover, we devise a new grasp generation model~\cite{ho2020denoising,huang2023diffusion}, enabling the efficient generation of pre-grasp poses for novel object configurations. For grasp execution, we propose a dual-stage policy that integrates motion planning for reaching and a learned policy for lifting. In simulations, our method demonstrates a success rate of $44.13\%$ in dual-object grasping with a Shadow Hand and shows scalability to handle more objects. Additionally, real-world experiments validate its practical effectiveness.

To summarize, the primary contributions of this work are:
(i) the creation of \dataset, a large-scale synthetic dataset explicitly curated for multi-object grasping research;
(ii) the formulation of the first \ac{gcrl} policy dedicated to the simultaneous grasping and lifting of multiple objects;
(iii) the augmentation of the execution policy for enhanced adaptability to novel object configurations and imprecise pre-grasp poses, achieved through specialist distillation and curriculum learning;
(iv) the comprehensive framework \method, which advances current robotic systems to achieve multi-object grasping.

\subsection{Related Work}

\paragraph*{Generating dexterous grasping}

Conditioned on the target objects, the generation of grasping poses for dexterous hands presents a complex challenge due to intricate kinematics and physical constraints. Research in this domain primarily bifurcates into analytical and data-driven approaches~\cite{bohg2013data}.

Analytical methods have a long history, with early research focusing on algorithmic solutions for hand- and object-specific grasping poses~\cite{ponce1993characterizing,ponce1997computing,li2003computing,eppner2021acronym}. GraspIt!~\cite{miller2004graspit} extended this to arbitrary hands and objects, albeit predominantly through a reach-and-squeeze strategy, limiting the diversity of grasping poses. Recent works have introduced generalized grasp quality metrics, such as force closure and \(Q_1\), enabling faster and more adaptable grasp synthesis~\cite{liu2021synthesizing,wang2023dexgraspnet,turpin2022graspd}.

Conversely, data-driven approaches learn the distribution of feasible grasping poses~\cite{jiang2021hand,lundell2021multi,huang2023diffusion}, or utilize proxy representations such as contact points~\cite{shao2020unigrasp} and contact maps~\cite{li2023gendexgrasp,xu2023unidexgrasp,brahmbhatt2019contactgrasp,grady2021contactopt}, conditioned on the characteristics of the objects.

Contemporary research endeavors to reach human-level interaction capabilities, including functional grasping~\cite{agarwal2023dexterous}, generalizable grasping~\cite{li2023gendexgrasp}, and multi-object grasping~\cite{yao2023exploiting}.

\paragraph*{Multi-object grasping}

The goal of multi-object grasping is to identify optimal hand configurations for holding multiple objects simultaneously. Research in this area predominantly follows two distinct methodologies. The first approach emphasizes grasping a collection of simple objects, such as balls, bricks, or pencils, with a focus on efficiency in the grasping process. This method often relies on the contacts between objects for grasp stability~\cite{yamada2015static,donald2000distributed,harada1998kinematics,agboh2022multi,sun2022multi}. It typically does not necessitate extensive kinematic redundancy, offering efficient grasping capabilities but at the cost of limited individual object manipulation. The second approach, however, utilizes the hand's kinematic redundancy by engaging different hand regions to grasp each object. This method allows for more detailed control over individual objects~\cite{yao2023exploiting}. Our work is more in line with this second approach, where the focus is on maintaining the ability to maneuver each object independently while also enhancing overall grasp efficiency.

\paragraph*{Reinforcement learning}

Robotic operation in complex physical environments often presents significant challenges for analytical solutions, particularly due to noisy sensory inputs. In such scenarios, \ac{rl} has become a favored choice for decision-making and control~\cite{hwangbo2019learning,lee2020learning,miki2022learning,jenelten2024dtc}.
The emphasis of \ac{rl}-based manipulation has largely been on single-object interactions~\cite{chen2022system,geng2023rlafford,xu2023unidexgrasp,wan2023unidexgrasp++,chen2022towards}. These approaches tend to be inadequate for learning the intricate grasping techniques necessary for dexterous hands.
For the complex task of multi-object grasping using dexterous hands where diverse object configurations pose additional challenges, conventional \ac{rl} strategies are often insufficient. In our work, we employ \ac{gcrl}~\cite{liu2022goal} to develop robust lifting policies, accelerated by IsaacGym~\cite{makoviychuk2021isaac}.

\subsection{Overview of \method}

Formally, we consider a tabletop scenario populated with multiple objects, denoted as \(\mathbf{O} = \{ O_j \}_{j=1}^{N_o}\). Each object \(O_j\) is represented as a point cloud in \(\mathbb{R}^{N \times 3}\), sampled from its surface \(S(O_j)\). The objective is to identify a sequence of hand actions, \(\mathcal{A} = \{a^t\}_{t=1}^{T}\), enabling the robotic hand to simultaneously grasp all objects. We primarily concentrate on scenarios where the objects are sufficiently proximate for simultaneous grasping.

\cref{fig:pipeline} illustrates the framework. For the given objects \(\mathbf{O}\), a pre-grasp pose \(H = (p, R, q)\) is proposed, encapsulating all targets; \(p\) denotes the hand's position, \(R\) its orientation, and \(q\) the joint angles. Two methods sample \(H\): a detailed synthetic algorithm (\cref{sec:synthesis}) and a fast generative model (\cref{sec:gen-grasps}). The grasp execution involves following a trajectory to \(H\) and then lifting the objects with a learned policy (\cref{sec:grasp-exe}).

\section{Pre-Grasp Pose Generation}

\subsection{Preliminaries}\label{sec:synthesis}

\begin{figure}[b!]
    \centering
    \includegraphics[width=\linewidth]{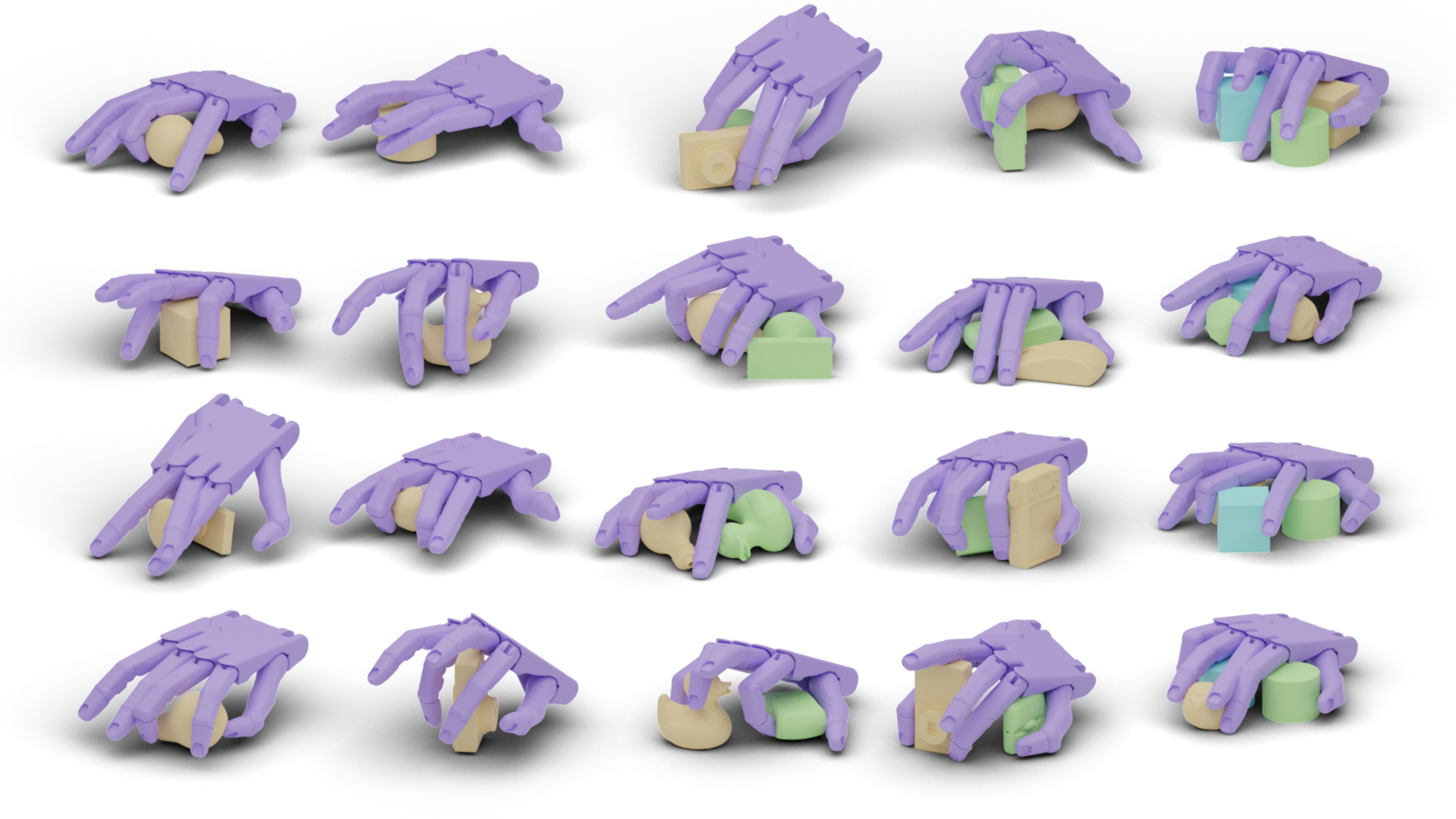}
    \caption{\textbf{Synthetic grasps using the augtmented \ac{dfc}.} From left to right: one (cols. 1-2), two (cols. 3-4), and three objects (col. 5).}
    \label{fig:multidfc-showcase}
\end{figure}

In multi-object grasping, a well-crafted pre-grasp pose is essential to meet the static force-closure conditions, serving as a goal for dynamic grasp execution. Leveraging the \ac{dfc} algorithm~\cite{liu2021synthesizing}, we generate diverse and stable pre-grasp poses for multiple objects. The hand configuration \(H\) is derived from the Gibbs distribution
\(
    p(H \vert \mathbf{O}) = \frac{p(H, \mathbf{O})}{p(\mathbf{O})} \propto p(H, \mathbf{O}) \sim \frac{1}{Z} e^{- E(H, \mathbf{O})},
\)
where the energy function \(E(H, \mathbf{O})\) integrates various terms:
\begin{equation}
    \small
    \begin{aligned}
        E(H, \mathbf{O})
        &= \sum_{j=1}^{N_o} \min_{x_j\subset S(H)} E_\mathrm{FC} (x_j, O_j) \\
        &+ \lambda_\mathrm{p} E_\mathrm{p}(H, \mathbf{O}) + \lambda_\mathrm{sp} E_\mathrm{sp}(H) + \lambda_\mathrm{q} E_\mathrm{q} (H),
    \label{eq:multi-dfc-energy}
    \end{aligned}
\end{equation}
where \(E_\mathrm{FC} (x_j, O_j)\) computes the force-closure error for object \(O_j\), with \(x_j\) representing the contact points on the hand that minimize this error. \(E_\mathrm{p}(H, \mathbf{O})\) penalizes any penetration between the hand and each object. Regarding the hand's configuration, \( E_\mathrm{sp}\) minimizes the self-collision of the hand, and \(E_\mathrm{q} (H)\) penalizes deviations of joint angles from their limits. The weights \(\lambda_{(\cdot)}\) are employed to balance these diverse energy components.
A gradient-based approach supplemented by the \ac{mala} optimizes \cref{eq:multi-dfc-energy} to avoid suboptimal local minima. The optimization process is parallelized across multiple initial states for efficiency. A filtering step is further applied to eliminate cases exceeding a predefined threshold. \cref{fig:multidfc-showcase} showcases the results of this synthesis for various object counts.
For algorithmic details and force-closure estimation, we direct readers to \cite{liu2021synthesizing}.

\subsection{Multi-object Grasp Generation}\label{sec:gen-grasps}

Using \ac{dfc} for multi-object pre-grasp pose generation is computationally demanding. To bypass this intensive optimization, we employ diffusion models, a family of generative models recently shown to be effective for complex data~\cite{urain2023se,huang2023diffusion}. Following \ac{ddpm}~\cite{ho2020denoising}, given object point clouds \(\mathbf{O}\), the grasp pose \(H = H^{(0)}\) is sampled through a denoising process:
\begin{equation}
    \small
    p(H^{(0)} \vert \mathbf{O}) = \prod_{t=1}^T p(H^{(t-1)} \vert H^{(t)}, \mathbf{O}),
\end{equation}
where
\begin{equation}
    \small
    \begin{aligned}
        &p(H^{(t-1)} \vert H^{(t)}, \mathbf{O}) = \mathcal{N} \left( H^{(t-1)}; \mu^{(t-1)}, \Sigma^{(t-1)} \right), \\
        &\mu^{(t-1)} = \mu_\theta (H^{(t)}, t, f_\theta (\mathbf{O})), \quad{}
        \Sigma^{(t-1)} = \Sigma (t).
    \end{aligned}
    \label{eq:grasp-diffuser-denoising-process}
\end{equation}

We adopt SceneDiffuser~\cite{huang2023diffusion} to learn \cref{eq:grasp-diffuser-denoising-process}. This model takes object point clouds as input and proposes a pre-grasp pose. PointNet++~\cite{qi2017pointnet++} is employed to extract feature vectors from each object's point cloud. These features, aggregated as \(N_o \times N_\mathrm{feat}\), act as the object conditions \(f_\theta (\mathbf{O})\). Cross-attention, computed with \(H^{(t)}\) as queries and object conditions as keys and values, is utilized in each sampling step.

Differentiating features from distinct objects, we append a learnable embedding to each feature vector, with identical embeddings for the same object and different ones for varied objects. To support part-level interaction reasoning between finger links and objects, 
drawing inspiration from several studies~\cite{zhang2021we,simeonov2022neural,seita2023toolflownet}, the hand configuration is represented by 31 keypoints on its links in Cartesian space, rather than joint angles in joint space. An optimization-based \ac{ik} solver derives joint angles from these keypoints.

For model training, we generated \dataset using our synthesis algorithm. The dataset for multi-object grasping includes $\approx{}90k$ synthetic pre-grasp poses, with a mix of $16.4k$ single- and $73.7k$ dual-object grasps featuring 8 objects (36 combinations) from YCB~\cite{calli2017yale} and ContactDB~\cite{brahmbhatt2019contactdb}. Objects are rescaled so that multiple ones can be fitted in one hand. Object pairs are randomly placed on a table with stable positions and orientations.

\begin{figure}[t!]
    \centering
    \includegraphics[width=\linewidth]{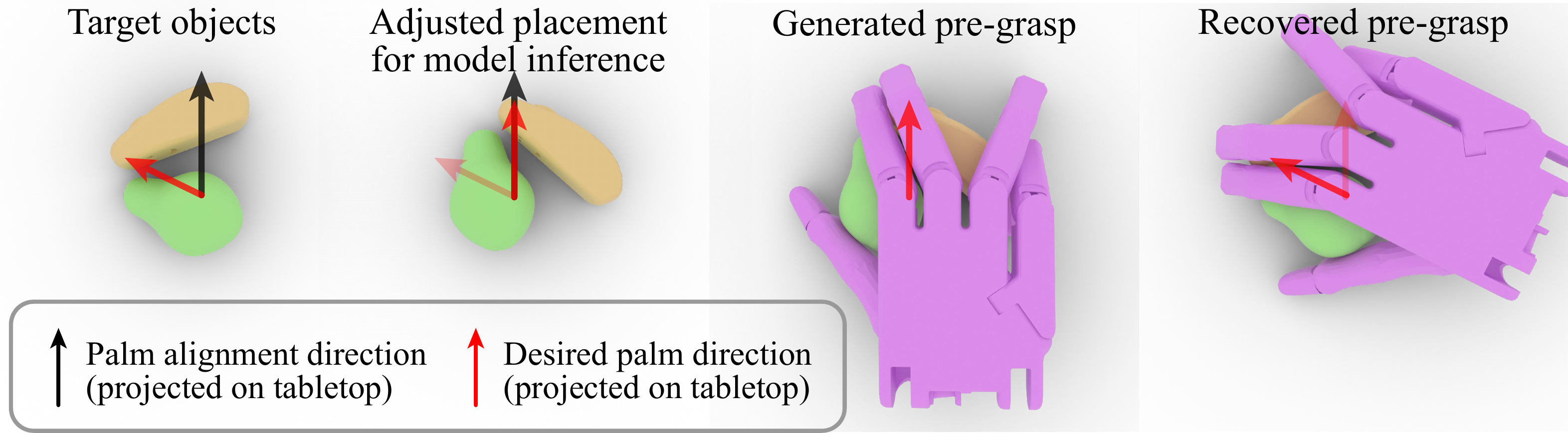}
    \caption{\textbf{Generate grasps with desired palm direction.} Given the desired palm direction (red arrow), the objects are rotated around the z-axis to align with the desired palm direction (black arrow). The pre-grasp pose is generated on the adjusted objects and is rotated to recover the pre-grasp under the initial object placement.}
    \label{fig:palm-align}
\end{figure}

While training, the data is preprocessed to align the palm direction (projected on the tabletop) by rotating the hand and objects around the z-axis, which simplifies learning by reducing one \ac{dof}. The model learns to generate palm-aligned hand poses, allowing control over the palm directions. This is done by rotating the target objects around the z-axis to align the desired palm direction with the alignment direction, and inverse the rotation on the generated hand pose to obtain the final hand poses (\cref{fig:palm-align}).

\subsection{Multi-object Grasp Refinement}\label{sec:method:refinement}

While the diffusion model shows potential in data generation, it occasionally results in imperfect grasps with penetration objects or insufficient contact. To rectify these shortcomings, we refine the hand configuration with optimization:
\begin{equation}
    \small
    \begin{aligned}
        \min_H E_g (H, \mathbf{O})
        & = E_\mathrm{p}(H, \mathbf{O}) \\
        & - \frac{\lambda_c}{N_o \vert S(H) \vert} \sum_{j=1}^{N_o} \sum_{\substack{x \in S(H)\\d(x, O_j) \leq \tau}} d(x, O_j),
    \end{aligned}
\end{equation}
where the first term is as defined in \cref{eq:multi-dfc-energy}, and the second term guides floating fingers closer to the surfaces of nearby objects for contact. Here, \(d(\cdot)\) is the distance from a potential contact point \(x\) on the hand to the surface of the object \(O_j\). To achieve a coarse-to-fine refinement, we linearly decrease the threshold \(\tau\) from \(2.0\mathrm{mm}\) to \(1.0\mathrm{mm}\) during optimization.

\section{Multi-object Grasp Execution}\label{sec:grasp-exe}

We execute the grasp with two phases: reaching and lifting (see \cref{fig:pipeline}C-E). The reaching phase involves motion planning, whereas the lifting phase utilizes a learned \ac{rl} policy.

\paragraph*{Reaching}

To plan a collision-free trajectory from the initial hand pose to the pre-grasp pose, we first compute their linear interpolation as a preliminary trajectory. This trajectory is further optimized to remove any hand-object penetrations at each timestep while ensuring a smooth temporal transition. Of note, while the grasp refinement (\cref{sec:method:refinement}) significantly reduces penetration, minor occurrences may still arise. However, the reaching phase could effectively compensate for such slight penetrations by causing minor object displacements. These disturbances will be accommodated by adapting the lifting policy (detailed in \cref{sec:specialists-to-generalist}).

\paragraph*{Lifting}

In real-world applications, conventional execution strategies like directly lifting from pre-grasp poses~\cite{miller2004graspit,she2022learning}, or squeezing fingers toward the closest object surface before lifting~\cite{li2023gendexgrasp}, often prove inadequate for multi-object grasping due to the complex nature of hand-object contacts. To address this, we employ \ac{gcrl} for precise and adaptable control over the intricate dynamics involved in hand-object interactions.

\subsection{Learning a Multi-Object Lifting Policy}

For our lifting policy, we utilize \ac{ppo}~\cite{schulman2017proximal} for learning within a simulated environment. Beginning from the pre-grasp pose of both the hands and objects, this policy controls the hand's pose and joint angles to lift all objects, guided by a reward:
\begin{equation}
    r = \omega_\mathrm{lift} r_\mathrm{lift} + \omega_\mathrm{succ} \mathbf{1}_\mathrm{succ} + \omega_\mathrm{r} r_\mathrm{r} + \omega_\mathrm{q} r_\mathrm{q} + \omega_\mathrm{obj} r_\mathrm{obj}, \label{eq:rl-rewards}
\end{equation}
\(r_\mathrm{lift} = \min_j h_j\) provides a dense reward based on object elevation, where \(h_j\) indicates the height of the \(j\)-th object). A success bonus (\(r_\mathrm{succ}\)) is awarded for elevating all objects above \(15 \mathrm{cm}\). We also find it beneficial to include rewards for maintaining goal hand rotation (\(r_\mathrm{r}\)), joint angles (\(r_\mathrm{q}\)), and object positions relative to the hand (\(r_\mathrm{obj}\)). These rewards are visualized in \cref{fig:pipeline}D.

The observation space for our policy is outlined in \cref{tab:obs-act-space}(a). It incorporates the current and goal states of both the hand and objects, along with their respective residues. Geometric information is sourced from hand and object features, extracted from point clouds using a pre-trained PointNet~\cite{qi2017pointnet}, all within the palm's coordinate frame. To optimize sample efficiency, the policy is trained across 512 parallel environments in IsaacGym~\cite{makoviychuk2021isaac}, each with a unique pre-grasp pose from \dataset's training set. We periodically refresh these poses to ensure comprehensive dataset coverage. The entire training process spans 8000 iterations, approximately taking 4 hours on a single NVIDIA A100.

In real-world scenarios, only the hand's state and the point cloud captured by depth cameras are typically available. To adapt the policy for such conditions, we distill it into a vision-based version using DAgger~\cite{ross2011reduction}. This involves replacing object features and states in the observation space (marked with $*$ in \cref{tab:obs-act-space}) with PointNet features of the hand-object scene, captured by three RGB-D cameras. The point cloud is also re-sampled to ensure even point distribution.

\begin{table}[b!]
    \centering
    \small
    \caption{\textbf{Observation and action definitions of our lifting policy.} The policy's 12-dimensional \textit{state} encompasses the position, orientation (represented by XYZ Euler angles), linear velocity, and angular velocity. The \textit{residue} is calculated as the difference between the current and goal values. Markers $^*$ and $^{**}$ indicate elements specific to the state-based and vision-based policies, respectively.}
    \begin{subtable}{\linewidth}
        \centering
        \caption{Observation space.}
        \begin{tabular}{cc}
            \toprule
                \textbf{Observation} & \textbf{Dimensions} \\
            \midrule
                Hand joint angles, velocities, forces & \(22 \times 3\) \\
                Hand fingertip wrenches & \(5 \times 6\) \\
                Hand base state & 12 \\
                Last actions & 24 \\
            \midrule
                Hand joint angle goals and residues & 22 + 22 \\
                Hand orientation goal and residue & 3 + 3 \\
                Hand point cloud feature & 256 \\
            \midrule
                Object \(O_i\) state $^*$ & 12 \\
                Object \(O_i\) pose goal and residue $^*$ & 6 + 6 \\
                Object \(O_i\) point cloud feature $^*$ & 256 \\
            \midrule
                Scene point cloud feature $^{**}$ & 256 \\
            \bottomrule
        \end{tabular}
    \end{subtable}%
    \\%
    \begin{subtable}{\linewidth}
        \centering
        \vspace{6pt}
        \caption{Action space.}
        \begin{tabular}{cc}
            \toprule
                \textbf{Action} & \textbf{Dimensions} \\
            \midrule
                Hand joint angle targets (actuated) & 18 \\
                Hand base wrench & 6 \\
            \bottomrule
        \end{tabular}
    \end{subtable}
    \label{tab:obs-act-space}
\end{table}

\subsection{Learning a Generalist from Specialists}\label{sec:specialists-to-generalist}

We find that the lifting policy's efficacy varies with object configurations and pre-grasp poses; \eg, when handling spheres versus cylinders or objects placed in various poses. To develop a generalist policy capable of adapting to various scenarios, we draw inspiration from previous works \cite{wan2023unidexgrasp++,zakka2023robopianist,lee2020learning,chen2020learning}. While retaining the settings from \cref{sec:grasp-exe}, we categorize the grasp data into distinct clusters based on object combinations and their placements relative to the palm. A dedicated specialist policy is trained for each cluster to master the grasps within, which provides demonstrations in distilling the vision-based generalist policy.

\subsection{Adapting to Imprecise Pre-Grasp Poses}\label{sec:imprecise}

Even with refinement, the generated grasps may be imperfect, and the execution of the planned reaching trajectories might inadvertently cause collisions that displaces objects. Policies trained exclusively on high-quality synthetic data are ill-suited for such scenarios, and training directly on these poses is ineffective due to the distorted goal representation practically. To address this challenge, we implement a structured learning curriculum. First, we generate imprecise poses by using the trained \ac{ddpm} on randomly positioned objects. Critically, we capture the states of the hand and objects at the end of the reaching phase, including moments where fingers may have shifted the objects. Next, the training regimen is divided into three phases: the first phase focuses on synthetic pre-grasp poses; the second phase introduces these imprecise poses, encompassing both synthetic and generated instances where objects have been displaced; the final phase predominantly incorporates generated data with moved objects.

It is crucial to highlight that, while our approach builds upon existing methodologies~\cite{jiang2021hand,lundell2021multi,huang2023diffusion,geng2023rlafford,xu2023unidexgrasp,wan2023unidexgrasp++,chen2022towards}, our concentration on dexterous robotic hands differentiates it from the referenced studies. The complexities involved in designing and controlling high-\ac{dof} dexterous robotic hands are distinct and more intricate. This necessitates substantial modifications and enhancements to the established methods, aligning them with the unique demands of dexterous manipulation.

\section{Simulations and Experiments}

In accordance with the evaluation protocol detailed in \cref{sec:exp:protocol}, we conduct a comprehensive validation of the proposed method. Quantitative results pertaining to the pre-grasp poses are elaborated in \cref{sec:exp:pregrasp}. The execution phase of the method is assessed in \cref{sec:exp:execution}, and ablation studies that examine the impact of various components of our approach are presented in \cref{sec:exp:ablation}. To demonstrate the method's ability to generalize, we provide results from simulations involving multiple objects and from real-world tests with dual-object setups (\cref{sec:exp:results}). We finally discuss observed failure cases and their implications.

\subsection{Evaluation Protocols}\label{sec:exp:protocol}

We generate random table-top object arrangements for each multi-object combination, following the same procedure as in our dataset (\cref{sec:gen-grasps}). For each arrangement, we generate pre-grasp poses, discarding those with significant penetration or insufficient contact ratio. In the execution phase, a grasp is deemed successful if all objects are elevated above $10 \mathrm{cm}$. The policy is evaluated on 512 unique poses, with five trials per pose, to compute the average success rate.

For dual-object grasping, the 8 objects in our dataset result in $C_8^1 + C_8^2 = 36$ unique combinations. We designate 8 pairs as unseen combinations, covering all objects, to evaluate our framework's generalization capabilities. Further, 6 combinations with 3 out-of-domain objects are introduced to challenge the framework with unfamiliar geometries.
To cluster the grasps into bins (\cref{sec:specialists-to-generalist}), we first group them by object combination and then further subdivide them based on the direction line connecting object centers in the palm's frame, resulting in 6 bins. We observe that grasping becomes particularly challenging when objects are aligned parallel to the forearm, primarily due to difficulties in achieving secure force closure. To mitigate this, we utilize the palm-alignment strategy from \cref{sec:gen-grasps} to limit palm directions in generated pre-grasp poses, thereby avoiding challenging configurations.

For comparison, we select three dexterous grasping methods, despite none being directly comparable to our task. We first contrast with a vision-only variant of our framework. GenDexGrasp~\cite{li2023gendexgrasp}, initially designed for single-object grasping using contact maps, is extended to multi-object scenarios with \dataset and evaluated using our execution framework. IBS-Grasping~\cite{she2022learning} learns single-object grasps considering the \ac{ibs} with \ac{rl}. We adapt it by integrating \dataset as initial off-policy demonstrations and modifying the reward structure for multi-object grasping.

The baselines are tested with pre-grasp poses generated for the same set of unseen object placements. We assess the effectiveness of these poses by training a single state-based policy for each method. For our approach and GenDexGrasp, we implement a state-based policy as outlined in \cref{sec:grasp-exe}. For \ac{ibs}-Grasping, we adhere to its original approach, directly lifting objects from the generated pre-grasp poses.

\begin{table}[b!]
    \centering
    \small
    \caption{\textbf{Quantitative evaluations on our method and adapted baselines~\cite{li2023gendexgrasp,she2022learning}.}
    Abbreviations are defined in \cref{sec:exp:pregrasp}. Success rates for specialist and generalist approaches are denoted separately, divided by ``/''. Notably, the performance of the baseline marked with $^{*}$ was evaluated using its original implementation~\cite{she2022learning} within the PyBullet environment.}
    \label{tab:results}
    \centering
    \label{tab:result}
    \resizebox{\linewidth}{!}{%
        \begin{tabular}{l c c c c c}
        \toprule
        \multirow{2}{*}{\textbf{Setting}} & \multicolumn{4}{c}{\textbf{Pre-Grasp Pose}} & \multicolumn{1}{c}{\textbf{Execution}} \\
        \cmidrule(lr){2-5}\cmidrule(lr){6-6}
         & $Q_1$ $\uparrow$ & PN $\downarrow$ & Div $\rightarrow$ & Time $\downarrow$ & Succ (\%) \\
        \midrule
        \textbf{\texttt{Syn-Pl}} & \textbf{0.30} & \textbf{1.64} & 8.54 & $\approx$ 1000 & \textbf{68.34} / \textbf{44.13} \\
        \textbf{\texttt{Syn-Com}} & \textbf{0.31} & 1.78 & 8.58 & $\approx$ 1000  & 26.73 \\
        \textbf{\texttt{Syn-Geo}} & \textbf{0.31} & 1.81 & 8.66 & $\approx$ 1000 & 22.55 \\
        \midrule
        \textbf{\texttt{Gen-Pl}} & 0.29 & \textbf{1.67} & 9.24 & \textbf{12.28} & 40.20 / \textbf{30.24} \\
        \textbf{\texttt{Gen-Com}} & 0.25 & 2.65 & 8.45 & 12.83 & 23.32 \\
        \textbf{\texttt{Gen-Geo}} & 0.29 & \textbf{1.54} & 9.14 & \textbf{12.08} & 15.65 \\
        \midrule
        \texttt{Syn-Pl-V} & \textbf{0.30} & \textbf{1.64} & 8.54 & $\approx$ 1000 & 0.59 \\
        \texttt{Gen-Pl-V} & 0.29 & \textbf{1.67} & 9.24 & \textbf{12.28} & 0.04 \\
        \midrule
        \texttt{GDG-Pl}~\cite{li2023gendexgrasp} & 0.27 & 27.75 & 4.23 & \multirowcell{3}{25.67\\{\scriptsize(32 samples)}} & 25.55 \\
        \texttt{GDG-Com}~\cite{li2023gendexgrasp} & \textbf{0.33} & 25.05 & 3.91 & & 14.57 \\
        \texttt{GDG-Geo}~\cite{li2023gendexgrasp} & 0.27 & 19.12 & 3.98 & & 6.91 \\
        \midrule
        \texttt{IBS-Pl}~\cite{she2022learning} & 0.23 & 36.29 & 7.09 & \multirowcell{3}{4.45\\{\scriptsize(1 sample)}} & 12.20$^{*}$ \\
        \texttt{IBS-Com}~\cite{she2022learning} & 0.22 & 35.92 & 7.40 & & 13.09$^{*}$ \\
        \texttt{IBS-Geo}~\cite{she2022learning} & 0.23 & 36.32 & 7.36 & & 14.18$^{*}$ \\
        \bottomrule
    \end{tabular}%
    }%
\end{table}

\subsection{Pre-Grasp Proposals}\label{sec:exp:pregrasp}

We assess the quality of synthesized or generated static poses using four metrics:
\begin{enumerate}
    \item \textbf{$\mathbf{Q_1}$ Metric}: This metric evaluates the largest radius of an origin-centered 6D sphere within the resistive wrench space~\cite{ferrari1992planning}, reflecting grasp stability. We compute \(Q_1\) for each object following Liu \etal~\cite{liu2020deep}. For multi-object grasps, we report the minimum $Q_1$ across all objects.
    \item \textbf{Penetration (PN)}: This is the maximal intersection depth (in \(\mathrm{mm}\)) between the hand, objects, and the table. Although a physically plausible grasp should be collision-free, slight penetrations are sometimes unavoidable due to numerical precision in computations.
    \item \textbf{Grasp Diversity (Div)}: This measures the average variance of joint angles (in \(\mathrm{deg}\)) across all grasp samples, indicating the range of grasping strategies. A versatile grasp generator should offer a variety of strategies.
    \item \textbf{Inference Time (Time)}: Measured in seconds on a single RTX 3090Ti GPU, this metric assesses efficiency. We test our method with a batch size of 256, GenDexGrasp with 32 to fit GPU memory capacity, and IBS-Grasping with 1 following its original implementation.
\end{enumerate}

Quantitative results in \cref{tab:results} highlight our method's proficiency in producing viable multi-object grasps efficiently. Synthetic grasps (\texttt{Syn}) provide high quality but are time-consuming, while generated grasps (\texttt{Gen}) balance quality and speed, maintaining satisfactory success rates. Remarkably, the generative model shows generalization to new object placements (\texttt{Gen-Pl}), combinations (\texttt{Gen-Com}), and geometries (\texttt{Gen-Geo}), although trained on only 8 objects. This is attributed to the diverse object configurations in the training data.
In comparison, GenDexGrasp~\cite{li2023gendexgrasp} (\texttt{GDG}) offers comparable quality but less diversity and tends to penetrate the table, indicating its limitation in tabletop scenarios. IBS-Grasping~\cite{she2022learning} (\texttt{IBS}) experiences more severe penetration, possibly due to its unstable stochastic policy.

\subsection{Execution Policy}\label{sec:exp:execution}

The execution phase is assessed in a simulator, with success defined as lifting all objects over \(10~\mathrm{cm}\). Each grasp undergoes five trials, and the average success rates are reported in \cref{tab:result}. Synthetic grasps (\texttt{Syn}) yield the highest success rate. For unseen placements, state-based specialists achieve an average success rate of \textbf{68.34\%}, while the vision-based generalist policy attains \textbf{44.13\%}. Despite their lower quality, generated grasps (\texttt{Gen}) still maintain reasonable success. However, success rates decrease for out-of-domain combinations (\texttt{-Com}) and geometries (\texttt{-Geo}).

During distillation, student policies show an approximate \(30\%\) decrease in success rates compared to their teachers, mainly due to two factors: (i) The vision-based generalist policy lacks direct access to object states, relying instead on scene observation through cameras, which leads to less precise observations. (ii) While each specialist focuses on a narrow set of similar grasping strategies, the generalist policy must adapt to a much wider range of pre-grasp poses and strategies. Although the performance drop, this policy is practical for real-world scenarios.

Comparatively, the baseline methods (\texttt{GDG} and \texttt{IBS}) demonstrate poorer performance in both grasp generation and execution across all generalization levels. Their limited grasp representations lack detailed kinematic information necessary for multi-object grasping. Furthermore, deep penetration during grasping leads to more collisions in reaching, resulting in significant object displacement. We also evaluated two visual baselines (\texttt{-Vis}), where a vision-based policy is learned directly from scratch using \ac{rl}. The coarse observation in the beginning of learning causes the failure in learning hand control, which highlights the importance of the distillation process.

\subsection{Ablations}\label{sec:exp:ablation}

\begin{table}[b!]
    \centering
    \small
    \vspace{-3pt}
    \caption{\textbf{Ablation studies.} Abbreviations are explained in \cref{sec:exp:pregrasp,sec:exp:ablation}. $^*$Evaluated on a subset of objects for efficiency.}
    \label{tab:ablation}
    \begin{subtable}{\linewidth}
        \centering
        \vspace{-2pt}
        \caption{Generative model}
        \label{tab:ablations:generator}
        \begin{tabular}{l c c c}
            \toprule
            \textbf{Setting}& $Q_1$ $\uparrow$ & PN $\downarrow$ & Succ (\%) \\
            \midrule
            \texttt{\textbf{Ours}} & \textbf{0.29} & \textbf{1.67} & \textbf{40.20} \\
            \texttt{Joint Angles} & 0.18 & 5.50 & 19.31 \\
            \texttt{w/o Obj Embd} & 0.27 & \textbf{1.38} & 37.21 \\
            \texttt{w/o Refinement} & 0.29 & 7.68 & 16.24 \\
            \bottomrule
        \end{tabular}
    \end{subtable}
    \hfill
    \begin{subtable}{0.47\linewidth}
        \centering
        \small
        \vspace{2pt}
        \caption{Specialist settings in \ac{rl}}
        \label{tab:ablations:generalists}
        \begin{tabular}{l c}
            \toprule
            \textbf{Setting} & Succ (\%) \\
            \midrule
            \texttt{\textbf{Ours}} & \textbf{40.20} \\
            \texttt{w/o Spe-Pl} & 29.09 \\
            \texttt{w/o Spe-Com} & 26.23 \\
            \texttt{w/o Spe} & 37.34 \\
            \bottomrule
        \end{tabular}
    \end{subtable}
    \hfill
    \begin{subtable}{0.47\linewidth}
        \centering
        \small
        \vspace{2pt}
        \caption{Training settings in \ac{rl}}        \label{tab:ablations:techniques}
        \begin{tabular}{l c}
            \toprule
            \textbf{Setting} & Succ (\%) \\
            \midrule
            \texttt{\textbf{Ours}$^*$} & \textbf{45.25} \\
            \texttt{w/o RL} & 1.37 \\
            \texttt{w/o Goal} & 16.79 \\
            \texttt{w/o Adpt} & 25.05 \\
            \texttt{w/o Curr} & 24.88 \\
            \bottomrule
        \end{tabular}
    \end{subtable}
    \hfill
\end{table}

We conduct ablation studies to dissect the components of our framework, focusing on grasp generation (\cref{tab:ablations:generator}) and execution policy (\cref{tab:ablations:generalists}). For brevity, we evaluate generated grasps on unseen object placements (\texttt{-Pl}), reporting only specialist performances which typically mirror student policy outcomes. We select four specialists covering all object placement bins to evaluate different execution techniques. These studies yield significant insights:

\paragraph*{Pre-grasp pose reasoning}

As discussed in \cref{sec:gen-grasps}, our model generates 31 keypoints to represent the hand. In contrast, directly generating joint angles (\texttt{Joint Angles}), a common approach in literature~\cite{jiang2021hand,xu2023unidexgrasp,wan2023unidexgrasp++}, resulted in decreased performance. This suggests that reasoning in Cartesian space, as done in our keypoint-based method, captures part-level hand-object interactions more effectively than joint space reasoning.
Additional ablations validate the impact of object embedding attachment (\texttt{w/o Obj Embd}) and the grasp refinement process (\texttt{w/o Refinement}).

\paragraph*{From specialists to generalists}

Our approach includes clustering the training data by object combinations and placements, training a specialist policy for each cluster. We explore variations including clustering solely by object combination (\texttt{w/o Spe-Pl}) or placement (\texttt{w/o Spe-Com}), and training a single policy for all grasps (\texttt{w/o Spe}). As per \cref{tab:ablations:generalists}, training specialists for specific placements and combinations marginally enhances expert demonstration quality. However, using specialists in isolation leads to reduced performance. While Xu \etal~\cite{xu2023unidexgrasp} highlighted the advantages of multiple specialists for varied object geometries, our dataset's limited diversity from 8 mostly convex objects might not necessitate distinct specialist policies. The true potential of specialists may emerge with more diverse object configurations.

\paragraph*{Training adaptive policy}

\cref{tab:ablations:techniques} emphasize the significance of our execution policy's design. The near-total failure of lifts without an \ac{rl} policy (\texttt{w/o RL}) underlines its critical role. Omitting observations and rewards for maintaining pre-grasp pose (\texttt{w/o Goal}) lowers the learning efficiency. Training exclusively on synthetic data without adapting to imprecise poses (\texttt{w/o Adpt}) or adapting without a structured curriculum (\texttt{w/o Curr}), results in suboptimal outcomes.

\begin{figure}[t!]
    \centering
    \includegraphics[width=\linewidth]{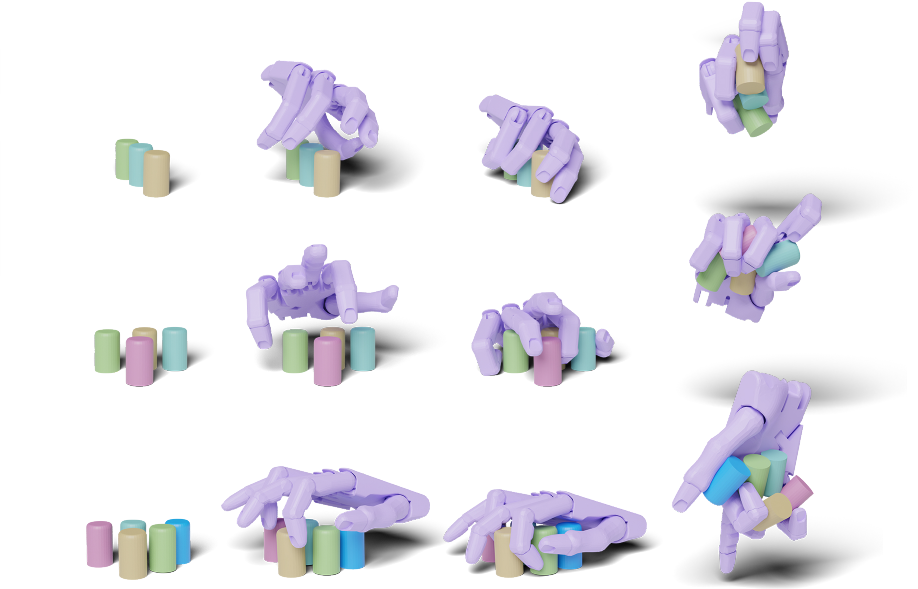}
    \caption{\textbf{Our framework supports grasping varying amounts (3-5) of objects.} Each row demonstrates the object placement and the execution process for different numbers of objects.}
    \label{fig:quantitative:object-amounts}
\end{figure}

\subsection{Additional Results}\label{sec:exp:results}

\paragraph*{Grasping more objects}

We test our framework's capability in grasping more than two objects. By synthesizing grasps for small cylinders and learning a state-based execution policy for each case, we assess the system's performance, as depicted in \cref{fig:quantitative:object-amounts}. The complexity of both pre-grasp proposal and execution escalates with the increase in object count. In scenarios with four objects, the hand's kinematic redundancy is increasingly utilized, and inter-object contact becomes essential for maintaining stability. When handling five objects, the hand needs to adopt an inverted position to effectively scoop the objects. These experiments underscore the scalability and probe the boundary of our approach.

\paragraph*{Real-world experiment}

\begin{figure}[t!]
    \centering
    \includegraphics[width=\linewidth]{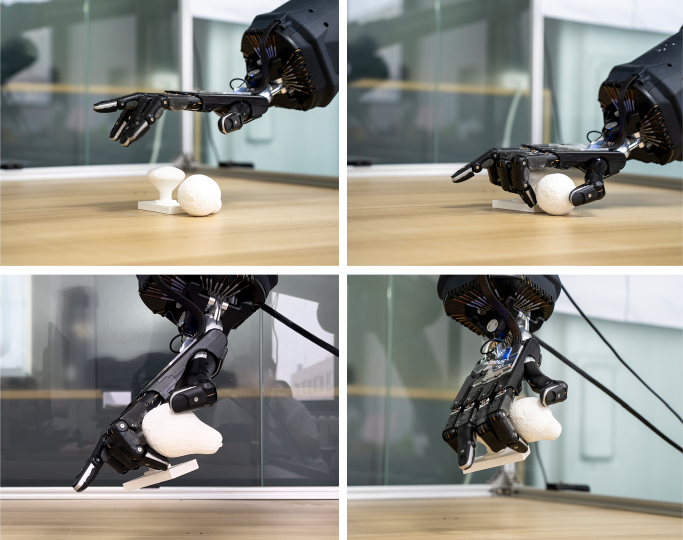}
    \caption{\textbf{Experiment with a Shadow Hand.} The sequence shows the phases of reaching, grasping, and lifting during the execution process.}
    \label{fig:real}
\end{figure}

Our method is further tested with a physical robot, using a Shadow Hand attached to a UR10e arm. Given the complexity of the task, we precompute execution trajectories in a simulation and then replicate them on the robot.
As illustrated in \cref{fig:real}, our approach successfully enables the robot to pick up two objects from a table, demonstrating its potential for real-world robotic applications.

\paragraph*{Failure cases}

\begin{figure}[t!]
    \centering
    \begin{subfigure}{0.45\linewidth}
        \includegraphics[width=\linewidth]{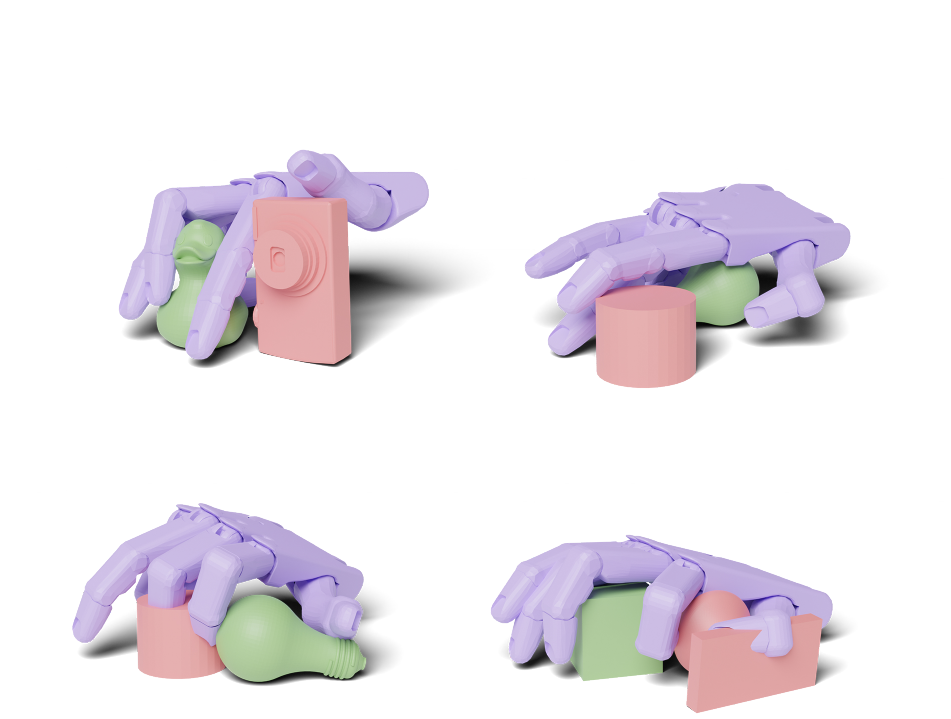}
        \caption{\textbf{Poorly generated samples:} Missing force-closure (top) and penetration (bottom).}
        \label{fig:failure:sample}
    \end{subfigure}%
    \hfill
    \begin{subfigure}{0.45\linewidth}
        \includegraphics[width=\linewidth]{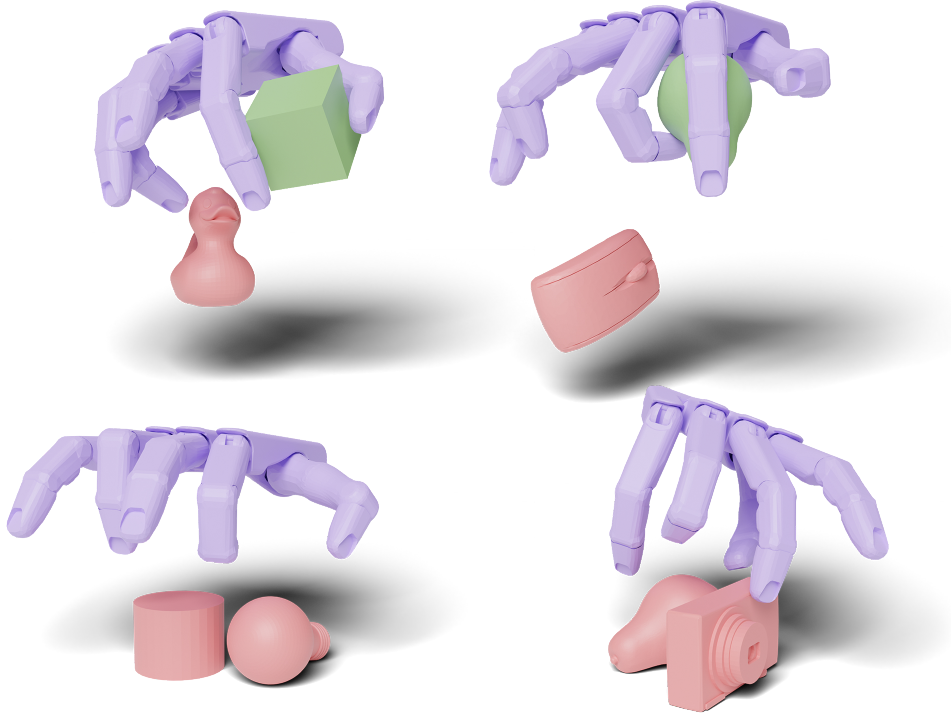}
        \caption{\textbf{Execution failures:} Dropping objects (top) and lift failures (bottom).}
        \label{fig:failure:execution}
    \end{subfigure}
    \caption{\textbf{Common grasp failures in generation and execution.} Objects marked in red indicate unsuccessful grasping attempts.}
    \label{fig:failure-cases}
\end{figure}

The primary causes are inadequate pre-grasp proposal quality and imprecise control during execution. \cref{fig:failure-cases} presents typical examples, including (i) poorly generated grasp samples, and (ii) instances during execution where objects are either are dropped mid-air or not lifted successfully.

\section{Discussions}

We introduced \method, a novel framework designed for the simultaneous grasping of multiple objects using multi-fingered robotic hands. Demonstrating the capability to manage various object counts, our approach underscores its potential for real-world implementations. This initiative lays the groundwork for future advancements in multi-object grasp planning and execution, aiming to boost the efficiency and adaptability of robotic grasping in real-world settings. Although our focus primarily lies on concurrent multi-object grasping, the feasibility of sequential object grasping, adopting an anthropomorphic strategy, is also acknowledged. Accordingly, we have refined our grasp synthesis algorithm to support efficient sequential grasping across different environmental conditions; please refer to the code repository.

The current study opens avenues for further exploration, particularly in narrowing the sim2real gap regarding perception and object dynamics, which could improve overall performance and enable more thorough real-world testing. Vision and proprioception, limited by significant occlusion and complex contact scenarios, hint at the potential enhancements tactile sensing could bring. Future research directions include exploring bimanual multi-object manipulation, where one hand stabilizes objects while the other performs precise tasks, such as insertion. Furthermore, our goal extends to equipping robots with capabilities for in-hand manipulation and tool use. Advancements in these domains aim to close the gap between robotic and human capabilities, empowering robots to undertake more sophisticated tasks and interactions.

\paragraph*{Acknowledgments}

We thank Qianxu Wang (PKU), Zihang Zhao (PKU), Junfeng Ni (THU), Nan Jiang (PKU), and Wanlin Li (BIGAI) for their helpful assistance.

\bibliographystyle{IEEEtran}
\balance
\bibliography{reference_header_shorter,reference}

\vfill

\end{document}